\documentclass[runningheads]{llncs}
\usepackage[T1]{fontenc}
\usepackage{graphicx}
\usepackage{subfigure}
\usepackage{booktabs} 
\usepackage{amsmath,amssymb,amsfonts}%

\begin{document}
\title{An Algorithm Board in Neural Decoding}
\titlerunning{Algorithm board}
%
\author{Jingyi Feng\inst{1} \and Kai Yang\inst{2,3}}
\authorrunning{Feng, and Yang.}
%
\institute{
School of Computer Science, Wuhan University \\
\and
School of Computer Science and Artificial Intelligence, Wuhan Textile University \\
\and
Hubei Luojia Laboratory, Wuhan\\
\email{fjy2035@gmail.com (Jingyi Feng); kyang@wtu.edu.cn (Kai Yang)}
}

\maketitle              
\begin{abstract}
Understanding the mechanisms of neural encoding and decoding has always been a highly interesting research topic in fields such as neuroscience and cognitive intelligence.
In prior studies, some researchers identified a symmetry in neural data decoded by unsupervised methods in motor scenarios and constructed a cognitive learning system based on this pattern (i.e., symmetry). Nevertheless, the distribution state of the data flow that significantly influences neural decoding positions still remains a mystery within the system, which further restricts the enhancement of the system's interpretability. Based on this, this paper mainly explores changes in the distribution state within the system from the machine learning and mathematical statistics perspectives.
In the experiment, we assessed the correctness of this symmetry using various tools and indicators commonly utilized in mathematics and statistics. According to the experimental results, the normal distribution (or Gaussian distribution) plays a crucial role in the decoding of prediction positions within the system. Eventually, an algorithm board similar to the Galton board was built to serve as the mathematical foundation of the discovered symmetry.

\end{abstract}
\section{Introduction}
\label{introduction}

At present, neural encoding is a functional model where neurons encode rich and intensive external stimuli into dynamic, time-varying neural activities, exploring the functional relationship between dynamic sensory stimuli and neural responses \cite{jia2023research}. Neural decoding is the restoration of external stimuli through a set of signals emitted by neurons, such as motion positions \cite{wallisch2014matlab}, image classification, and speech recognition.
Research on neural encoding and decoding has implications for our understanding of brain functioning mechanisms, the treatment of brain disorders, and the development of brain-computer interfaces and machine intelligence. 
Existing technologies mainly focus on movement, speech, and vision, aiming to scientifically understand the link between neural activity and the outside world \cite{livezey2021deep}. Moreover, prostheses, robots, mice, and other devices that fully realize "brain control technology" are becoming a reality \cite{mcfarland2008brain,kim2016commanding}.

Some researchers have achieved good results in decoding the monkey finger movement trajectory \cite{wallisch2014matlab} and the mouse movement trajectory \cite{Glaser2020machine} using supervised methods based on brain neural data, demonstrating the correspondence between the recorded neural data and external movements and their effectiveness.
Currently, there are related studies on the time-independence \cite{paranjape2019cross} and time-sequential \cite{wu2019neural} aspects of brain neural data, which have achieved good results. Furthermore, research on supervised methods \cite{wu2019neural} and unsupervised methods \cite{xue2017unsupervised} for neural decoding has also produced encouraging results.
In recent years, due to the rapid development of neural networks, deep learning methods have gradually played a positive role in neural decoding \cite{ahmadi2016decoding}. However, due to the unexplainability of deep learning, there are still significant shortcomings in exploring the inherent explanatory power of neural data, such as, the generation mechanism and intrinsic property of neural data.

Some studies still attempt to explore whether neural data have certain inherent properties from an unsupervised \cite{xue2017unsupervised} and weakly supervised perspective \cite{feng2018neural,feng2020weakly}. Among them, some researchers have found the symmetry between the unsupervised decoding positions and the ground-truth positions in the decoding of brain neural data for movement \cite{feng2018neural,feng2020weakly,feng2021vif,feng2023grid}. Here, "symmetry" refers to the unsupervised decoded trajectory showing a certain degree of symmetry with the actual trajectory within the activity space in the navigation task of brain neural data. Although some progress has been made based on this discovered symmetry as a basic hypothesis, the specific reason for this symmetry remains unknown. Based on this, in our paper, we are still exploring this issue, attempting to explore the key role that symmetry plays in decoding prediction within the system from common metrics in mathematical statistics.

The main contributions of this paper are summarized as follows:
\begin{itemize}
\item We have re-verified the existence of symmetry from a larger dataset of a rat's movement positions \cite{Glaser2020machine}, which is different from the dataset of a monkey's finger movement positions \cite{wallisch2014matlab}. We have also evaluated the symmetric data processing details from commonly used mathematical and statistical indicators.
\item We have evaluated decoding predictions and prediction errors (or noise) from probability density function (PDF) and power spectral density (PSD), and found that Gaussian distribution plays a crucial role within the system in decoding positions of this data.
\item We derived an algorithm board similar to the Galton board for serving as the mathematical foundation of the discovered symmetry.
\end{itemize}

The remainder of this paper is organized as follows.
Section 2 presents an overview of related work on neural decoding, including sequential supervised methods and a novel weakly supervised approach that has been discovered.
Section 3 describes the unsupervised method and correction approach adopted in this paper.
Section 4 experimentally validates the symmetry discovered in neural decoding and evaluates the decoding predictions and probability distribution that occur within the system while processing brain neural data from both quantitative and qualitative aspects in mathematical statistics.
Next, section 5 discusses and compares the similarities between an algorithm board derived from the discovered symmetry and the Galton board. Finally, section 6 briefly summarizes the main contributions of this study.

\textbf{Notable Contributions.} In previous studies, a discovered symmetry \cite{feng2018neural,feng2020weakly} in neural data decoded by unsupervised methods in motor scenarios, and built a cognitive learning system \cite{feng2021vif,feng2023grid} based on this pattern. However, the distribution state of data flow that has a significant impact on neural decoding positions remains a puzzle within the system. Finally, an algorithm board similar to the Galton board was constructed. From the perspective of classical mathematical statistics theory, specifically the Galton board, it represents the mathematical basis for the discovered symmetry.

\section{Related work}

In neural decoding, some studies often use supervised algorithms to achieve good decoding predictions, such as finger position decoding \cite{wallisch2014matlab,wu2019neural} and a rat's movement trajectory decoding \cite{Glaser2020machine}. Due to the temporal characteristics of neural decoding signals, temporal decoding algorithms have received much attention. Firstly, state-space models (SSMs) with associations between the current state and the previous state have become more developed \cite{gilja2012a,wallisch2014matlab,wu2019neural}. 
In addition, deep learning has been the focus of numerous studies, such as LSTM (Long Short-Term Memory) \cite{elango2016sequence,pan2018rapid}. Further, the source of the recorded neural activity can change from day to day, e.g., due to a slight movement of the implanted electrodes. The proposed multiplicative RNN (Recurrent Neural Network) allows mappings from the neural input to the motor output to partially change from neural activity \cite{sussillo2018making}. The decoding performed by these methods is more accurate than that of the traditional methods. 

In recent years, some researchers have proposed a novel weakly supervised method different from previous studies. This method is mainly based on a symmetric pattern discovered from decoding brain neural data between the unsupervised decoding positions and the ground-truth positions. This method has achieved decoding predictions that are much higher than those of unsupervised methods and close to those of supervised methods \cite{feng2018neural,feng2020weakly}. Based on these studies, a general framework for refining weakly supervised system has been proposed, which has been algorithmically validated \cite{feng2021vif} and theoretically justified from machine learning, neuroscience, cognitive science, and more \cite{feng2023grid}. Due to the complexity and difficulty in understanding this framework, further research is needed to explore the generation of symmetric mechanisms and the processing details of data in this system for the development of this framework. 

\section{Method}
\subsection{An introduction to an unsupervised method}

\begin{figure}   
    \centering
    \includegraphics[width=\textwidth]{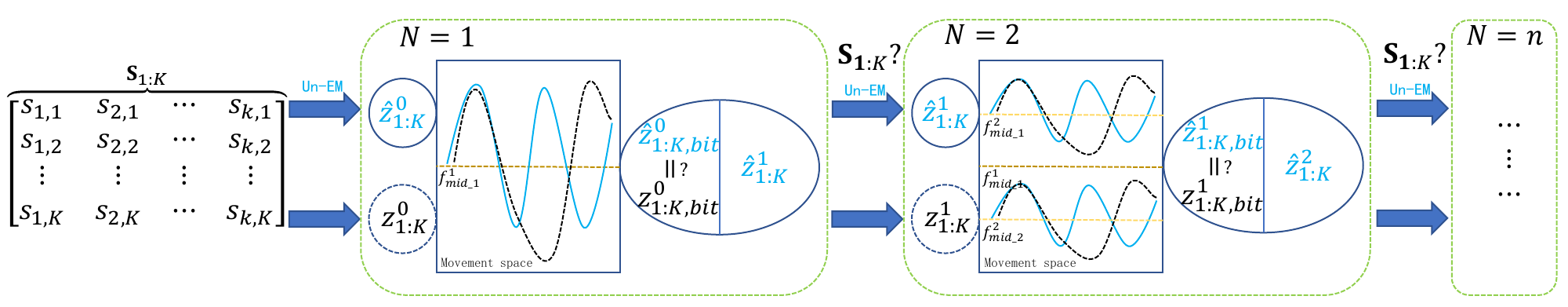}   
    \caption{First, neural data $\textbf{S}_k$ are decoded through an unsupervised EM algorithm to obtain predicted location $\hat{z}^0_{1:K}$. Then, this unsupervised predicted positions and the ground-truth trajectory ($z^0_{1:K}$) are approximately symmetric about the central axis ($f^1_{mid\_0}$) in the active space. Therefore, space is encoded as 0 and 1 above and below the central axis, respectively, which means that the ground-truth positions and predicted positions entering this upper or lower region are encoded as the corresponding 0 and 1. 
    Through analogy between predicted encoding values (0 or 1) and ground-truth encoding values (0 or 1), symmetrical correction obtains $\hat{z}^1_{1:K}$ when $N=1$. Here, $N$ is a spatial parameter to control the change from space to subspace. When $N=2$, similar unsupervised prediction and correction operations are performed in the corresponding subspace or sub-region, as in the case of $N=1$. This process is repeated successively. 
    }
    \label{fig: correction}
\end{figure}

In this paper, we mainly analyze the internal evaluation of data flow within the system, that is, the relationship between its unsupervised prediction and corrected prediction, as well as the ground-truth trajectory, as the $N$-value increases. 
However, in the external evaluation of system output, that is, the effect of the trained system model in testing, reference can be made to \cite {feng2018neural,feng2020weakly,feng2021vif}. They observe that the symmetry between predicted trajectory and ground-truth trajectory is mined by unsupervised algorithms, such as unsupervised KF (Kalman Filter) and unsupervised EM (Expectation Maximization) \cite{feng2020weakly}. 
In this paper, we focus on the crucial role of correction within the system and its impact on internal data transmission, evaluating it through quantitative and qualitative metrics. For demonstration purposes, unsupervised EM was adopted due to its excellent decoding performance and efficient runtime \cite{feng2020weakly}. 
In the iterative process of classical EM, the E-step and M-step are defined \cite{smith2005erratum} as a reference. 
The weight that considers the temporal correlation between any two Gaussian noise random variables at different times in the state space model is updated as \cite{feng2020weakly}:
\begin{eqnarray}
\ &\bf\tilde{W} &= {\left[\frac{K\sum_{k=1}^K{\bf{S}}_k\hat z_k - \sum_{k=1}^K\hat z_k\sum_{k=1}^K{\bf{S}}_k}{\sum_{k=1}^K\left(\hat z_k^2 + P_k\right) - \sum_{k=1}^K\hat z_k}, \frac{1}{K}\left(\sum\nolimits_{k=1}^K{\bf{S}}_k - {\bf\bar a}\sum\nolimits_{k=1}^K\hat z_k\right)\right]}^\mathrm{T}
\end{eqnarray}
where, $\bf\tilde{W}$ is the updated weight.
$\bf\bar{a}$ is equivalent to $\bf{a}$ in $\bf\tilde{W} = [\bf{a},\bullet]$.
$\hat z_k$ is the unsupervised predicted position or the predicted position after system correction.
${\bf{S}}_k$ are the input neural data.
$K$ is the data length.
$P_k$ is the covariance of $\hat z_k$ at time $k$.

\subsection{An introduction to a correction method}

Fig. \ref {fig: correction} shows the correction process from $N=0$ to $N=n$ within the system based on the symmetry discovered. Its main feature is the observation of a symmetry between the unsupervised predicted trajectory and the ground-truth trajectory in the active space, and then encoding the space with a bit-value (0 or 1) to correct the predicted trajectory through analogy. Experiments have shown that this method has strong decoding ability and robustness \cite{feng2020weakly,feng2021vif}. According to Fig. \ref{fig: correction}, the position encoding and position correction within the system are as follows:

\begin{equation}
    \hat{Z}^n_{k, bit}=\left\{
             \begin{array}{llr}
                1  \quad if \; \hat z^n_k \; or \; z^n_{k,bit} \; \geq f^n_{mid}(\bullet) \\
                0  \quad if \; \hat z^n_k \; or \; z^n_{k,bit} \; < f^n_{mid}(\bullet)
             \end{array}
    \right.
\end{equation}
\begin{equation}
    \hat{Z}^{n+1}_{k}=\left\{
             \begin{array}{llr}
                \hat z^n_{k}  & if \; \hat z^n_{k,bit}=z^n_{k,bit}  \\
                \hat z^n_{k} + 2(f^n_{mid}(\bullet) - \hat z^n_{k})  & if \; \hat z^n_{k,bit} \neq z^n_{k,bit} 
             \end{array}
    \right.
\end{equation}

Where, formula (2) is for encoding the external positions in the active space. 
$\hat{Z}^n_{k, bit}$ is the bit-value (0 or 1) encoded for the unsupervised predicted position $\hat z^n_k$ or the ground-truth position $z^n_k$ at the $k$-th moment or the $k$-th sample when $n=N-1$. 
$\hat z^n_k$ is the unsupervised predicted position. 
$f^n_{mid}(\bullet) = \frac{(z_{max} + z_{min})}{2}$ is the median value of the active space or subspace when $n=N$, where $z_{max}$ and $z_{min}$ are the maximum and minimum boundaries of movement, respectively. 
Formula (3) is for correcting the unsupervised predicted position in the active space. 
$\hat{Z}^{n+1}_{k}$ is the predicted position after correcting the unsupervised prediction position when $n=N$ by using the analogy between the encoded bit-values at the $k$-th time or the $k$-th sample. 
$\hat z^n_{k,bit}$ and $z^n_{k,bit}$ are the bit-values (0 or 1) encoded for the unsupervised prediction position and ground-truth position, respectively, namely $\hat{Z}^n_{k, bit}$. 

In summary, within the system and in terms of the entire data flow, correction plays a key role. The reason for the correction is that when unsupervised methods decode brain neural data, their unsupervised predicted trajectory and ground-truth trajectory exhibit symmetry in the active space. 
In previous studies \cite{feng2018neural,feng2020weakly,feng2021vif,feng2023grid}, although some researchers have discovered this symmetry, they have not addressed the mathematical and statistical depth of the interpretation of this symmetry within the system. These also limit the further development of this discovery from another perspective. 
In this paper, we mainly focus on these defects. The ultimate goal is to deeply explain what characteristics the data that generates this symmetry has, and whether these characteristics can provide us with new insights and promote the scientific growth of this discovery.

\section{Experiment}
\subsection{The dataset and evaluation metrics}
\label{data}

The data was collected from the Institutional Animal Care and Use Committees of the Approprity Institutions and can be found at https://github.com/ KordingLab/Neural$\_$Decoding \cite{Glaser2020machine}, which is larger than the dataset \cite{wallisch2014matlab}. 
From this recording \cite{Glaser2020machine}, 46 neurons are used over 75 minutes. These neurons had mean and median firing rates of 1.7 and 0.2 spikes/sec, respectively. 
Then, the dataset has 219,089 time points, which are transformed into 28039 samples after processing. 
A rat’s activity space is about length $L = 200cm$, and width $B = 200cm$. In \cite{Glaser2020machine}, some supervised algorithms, such as WienerFilter and WienerCascade, etc., are utilized to decode the neural data into the moving positions of a rat. In this paper, the dataset (70\%) is used as the training set.

The evaluation metrics can be calculated as follows. 
\textbf{$\boldsymbol{R^2}$:} $R^2 = 1 - \frac{\sum\nolimits_{k=0}^{K-1}(\hat{z}_k - z_k)^2}{\sum\nolimits_{k=0}^{K-1}(z_k - \Bar{z})^2}$. 
Where, $\hat{z}_k$ and $z_k$ are the decoded position and the ground-truth position in the $x$-axis or $y$-axis of the test data. $\Bar{z}$ is the mean value of the $\hat{z}_k$. $K$ is the total number of samples in the test data. 
R-squared, often abbreviated as $R^2$ or R-sq, is commonly used to measure the fitting degree of linear regression. 
$R^2=1$ indicates that the model predicts all ground-truth labels without bias, while $0<R^2<1$ indicates that the model's fitting level is better than the mean. $R^2 \leq 0$ indicates that the model's fitting level is close to or inferior to the mean, indicating that the model is of low research value.
\textbf{RMSE:} $RMSE=\sqrt{\frac{1}{K}{\sum\nolimits_{k=0}^{K-1}(\hat{z}_k-z_k)^2}}$. 
RMSE is the root mean square error. 
\textbf{PCC:} $PCC=\frac{\sum\nolimits_{k=0}^{K-1}(\hat{z}_k-\Bar{\hat{x}})(z_k-\Bar{z})}{\sqrt{\sum\nolimits_{k=0}^{K-1}(\hat{z}_k-\Bar{\hat{z}})^2(z_k-\Bar{z})^2}}$. 
Where, $\Bar{\hat{x}}$ is the mean value of the $\hat{x}$. 
PCC (Pearson correlation coefficient) is often used to measure the linear correlation between two variables, with a value close to $PCC=1$ indicating a strong positive correlation, a value close to $PCC=-1$ indicating a strong negative correlation, and a value close to $PCC=0$ indicating no significant relationship. 
\textbf{$\boldsymbol{R_{max}}$:} $R_{max}=\frac{L}{2^N}$\cite{feng2021vif}. 
Where, $R_{max} \geq 0$ is the mean length of each segment of a line segment after it has been cut, which is the maximum range of robustness. The higher the value, the stronger the robustness of the system, while the lower the value, the lower the robustness of the system. 
$L$ generally refers to the total length of a line segment in terms of length, width or height, etc. $N$ is the number of times a segment has been cut. 
\textbf{KL-score:} $KL\text{-score} (p, q) =\sum\nolimits_{k=0}^{K-1}p(\hat{z}_k)log\frac{p(\hat{z}_k)}{q(z_k)}$. 
Where, $p$ and $q$ respectively represent probability distribution $p$ and probability distribution $q$. 
KL (Kullback-Leibler) divergence is used to measure the distance or similarity between the probability distributions from which two variables are generated. 
$KL\text{-score} \geq 0$, the smaller the value, the closer the two probability distributions are, and 0 indicates a consistent probability distribution. 
\textbf{JS-score:} $JS\text{-score} (p, q) = \frac{1}{2}KL\text{-score} (p, \frac{p+q}{2}) + \frac{1}{2}KL\text{-score} (q, \frac{p+q}{2})$. 
The JS (Jensen-Shannon) divergence is an improvement over the KL divergence and a symmetrical way to measure the similarity between two probability distributions.

\subsection{The discovered symmetry}

\begin{figure}    
    \centering
    \subfigure[Ground-truth positions and predicted positions (N=0, N=1)]{
        \includegraphics[width=0.47\textwidth]{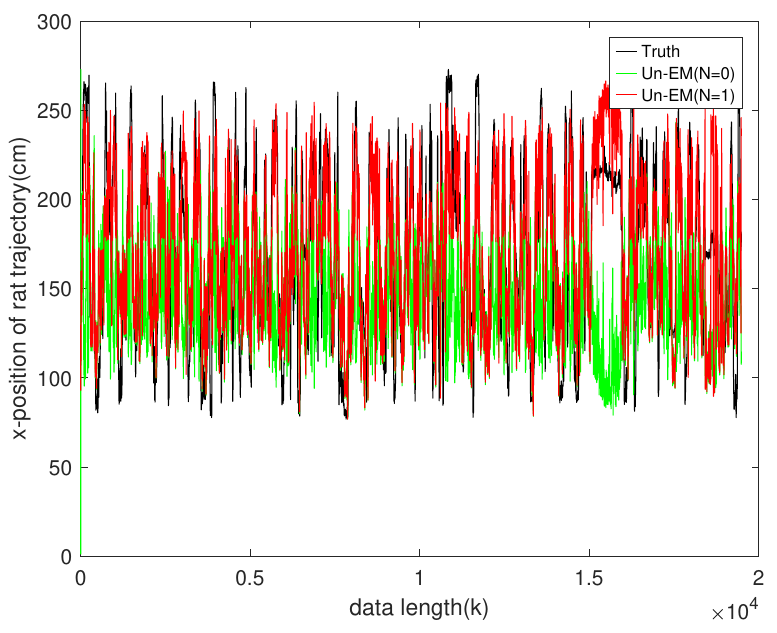}
    }
    \subfigure[3000 samples are extracted from the (a)]{
        \includegraphics[width=0.47\textwidth]{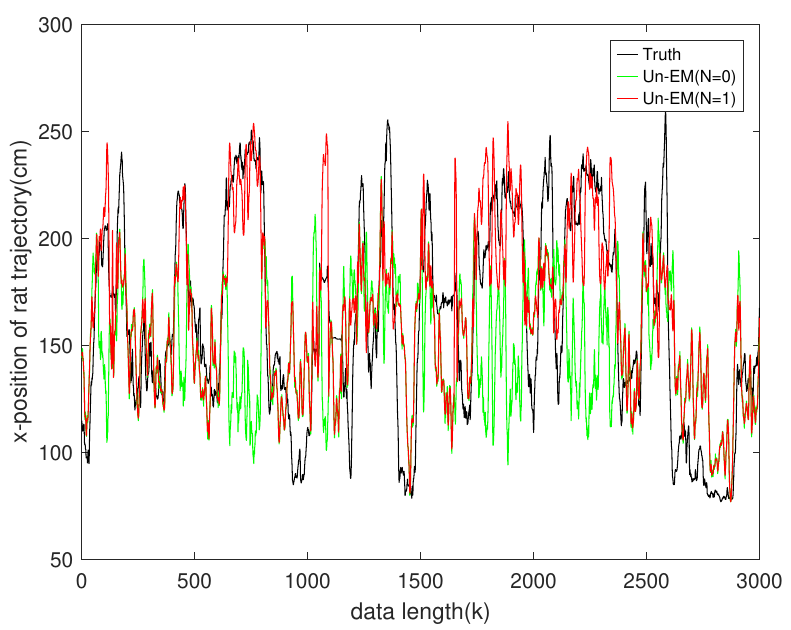}
    }
    \caption{(a) The figure shows the visualization of the unsupervised predicted trajectory (green line), the predicted trajectory after one correction (red line), and the ground-truth trajectory (black line). (b) The figure demonstrates the enlarged and visualized observation of approximately 3000 samples extracted from Fig. (a). As can be seen, the green trajectory undergoes symmetric correction to produce the red trajectory, which tracks the black trajectory almost perfectly. These indicate that the corrected predicted trajectory has high accuracy.
    }
    \label{fig: symmetry}
\end{figure}

Currently, in the datasets \cite{wallisch2014matlab,Glaser2020machine} we have tested, we have identified similar symmetrical properties in unsupervised decoding positions of brain neural data. In this paper, the dataset \cite{Glaser2020machine} is adopted to demonstrate and illustrate this symmetry, as shown in Fig. \ref{fig: symmetry}. 
Fig. \ref{fig: symmetry} displays the unsupervised predicted trajectory ($N=0$) and the corrected unsupervised predicted trajectory ($N=1$) visualized in $x$-position, along with the ground-true trajectory. The green line represents the unsupervised predicted trajectory, while the red line represents the corrected predicted trajectory. 
To provide a clearer visualization, Fig. \ref{fig: symmetry} (b) extracts approximately 3000 samples from Fig. \ref{fig: symmetry} (a). 
From the figure, it can be seen that the unsupervised predicted trajectory is roughly symmetrical about the midline of the activity space with the ground-truth trajectory. Based on this, the corrected predicted trajectory (red line) exhibits a basic agreement with the ground-true trajectory, indicating a high prediction accuracy after one correction. 
This further validates the existence of this symmetrical attribute.

\begin{figure}    
    \centering
    \subfigure[The scatter of the ground-truth and prediction (N=0,1,2)]{
        \includegraphics[width=0.47\textwidth]{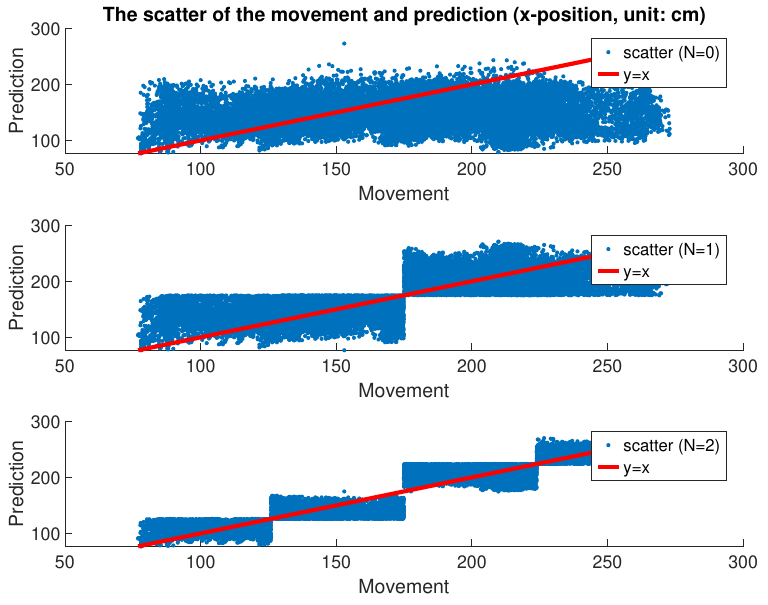}
    }
    \subfigure[The scatter of the ground-truth and prediction (N=3,4,5)]{
        \includegraphics[width=0.47\textwidth]{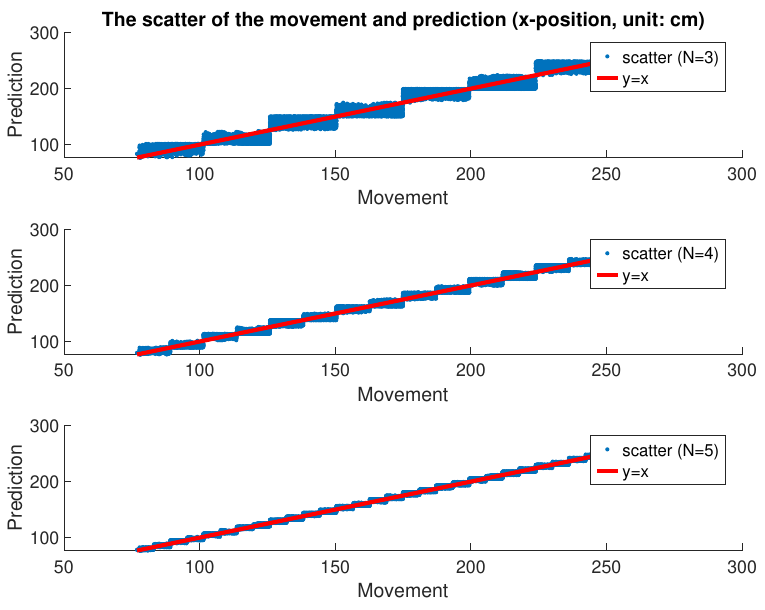}
    }
    \caption{(a) and (b) respectively show the scatterplot between the predicted positions and the ground-truth positions from $N=0$ to $N=2$, and from $N=3 $ to $N=5$. Where $X$-axis represents the ground-truth positions, and $y$-axis guarantees the unsupervised prediction positions ($N=0$) and the corrected prediction position ($N>0$).
    }
    \label{fig: scatter}
\end{figure}

Additionally, in order to show the role of the symmetry operation, Fig. \ref{fig: scatter} displays a scatterplot of the moving ground-truth positions and the unsupervised predicted positions ($N=0$), as well as the corrected predicted positions ($N>0$) in the $x$-positions. 
Among them, the $x$-axis represents the moving ground-truth positions, while the $y$-axis represents the unsupervised predicted positions and the corrected predicted positions. The red line segment represents the ground-truth positions equal to the predicted positions, i.e., $y=x$. 
The scatterplot provides us with the following insights:
(1) On both sides of the red line, the distribution of scatter points is relatively uniform, and this pattern persists as the $N$-value increases; 
(2) As the $N$-value increases, the predicted trajectory approaches the ground-truth trajectory in an exponential manner, i.e., $2^N$; 
(3) As the $N$-value increases, after each symmetrical correction, the predicted positions gradually approaches the ground-truth positions and fluctuates around it; 
(4) As the $N$-value becomes larger, with more corrections, the system's robustness to noise in the hidden space (such as the length and width of the rectangular box) decreases significantly; (5) Conversely, as the $N$-value becomes smaller, with fewer corrections, the system's robustness to noise is significantly higher.

\subsection{The quantitative metrics of the system in mathematics and statistics}

In this section, we evaluate the decoding prediction in system internal correction from different views using various commonly used metrics. These metrics help us fully grasp the changes in the system during neural data processing. 
Among them, $R^2$ and RMSE are used to assess the proximity of predicted and ground-truth positions, while PCC is used to evaluate whether the predicted and ground-truth trajectories are linearly related. KL-divergence and JS-divergence are used to evaluate whether the predicted and ground-truth positions have consistent distribution.

Table \ref{fig: N-value-x} shows the quantitative metrics for the evaluation of decoding predictions on $x$-position as the parameter $N$ increases. Among them, maximum robustness ($R_{max}$) will decrease as the $N$-value increases. 
When $N=0$, $R^2=-0.503$ indicates that the unsupervised prediction is below the mean value, and the unsupervised prediction is typically not considered useful. As observed in Fig. \ref{fig: scatter}, the scatterplot between predicted and ground-truth positions shows significant randomness when $N=0$. However, when $N>0$, the $R^2$ evaluation is significantly higher than the mean value, reaching 0.5, and approaching 1 for $N>2$. 
Additionally, RMSE provides more specific numerical values for prediction errors, which complements the $R^2$ metric. 
Then, in correlation coefficients (i.e., PCC), when $N=0$, PCC are respectively close to zero, indicating that the unsupervised predicted positions and the ground-truth positions are not linearly related. This corresponds to the scatterplot in Fig. \ref{fig: scatter}, which shows significant randomness when $N=0$. As $N>0$, the PCC can quickly approach 1, indicating a strong linear relationship between the predicted and ground-truth trajectories. 
In the analysis above, we consider that the PCC serves a similar purpose as $R^2$ because their numerical values are closer. Further, the numerical values of PCC appear to change more rapidly, such as when $N=2$, they reach a value of 0.7, which is higher than $R^2$'s 0.5. The same patterns can be observed in both Table \ref{fig: N-value-x} and Table \ref{fig: N-value-y}. 
Finally, we mainly focus on the changes in KL-divergence and JS-divergence, which are mainly used to evaluate the distribution consistency between two data vectors. If two data distributions are completely identical, the divergence value is zero-value. From both tables, we can see that as the $N$-value increases, the distribution of the predicted and ground-truth positions gradually converges toward consistency. For a more detailed discussion of this topic, please refer to the experimental observations on probability density distribution in section \ref{xy-positions}.

\begin{table}
\caption{x-position, the correction experiment within the system in a rat's trajectory.}
\label{fig: N-value-x}
\begin{tabular}{|l|l|l|l|l|l|l|l|l|}
\hline
Parameter & $R^2$ & RMSE & PCC & $R_{max}$ & KL-score ($\times 10^{-3}$) & JS-score ($\times 10^{-3}$) \\
\hline
$N=0$ & -0.503 & 57.937 & 0.096 & 200  & 53.527 & 13.327 \\
$N=1$ & 0.509  & 33.106 & 0.727 & 100  & 19.781 & 4.979 \\
$N=2$ & 0.837  & 19.076 & 0.919 & 50   & 6.716  & 1.675 \\
$N=3$ & 0.959  & 9.527  & 0.980 & 25   & 1.640  & 0.409 \\
$N=4$ & 0.990  & 4.840  & 0.995 & 12.5 & 0.432  & 0.108 \\
$N=5$ & 0.997  & 2.420  & 0.999 & 6.25 & 0.109  & 0.027 \\
\hline
\end{tabular}
\end{table}

\begin{table}
\caption{y-position, the correction experiment within the system in a rat's trajectory.}
\label{fig: N-value-y}
\begin{tabular}{|l|l|l|l|l|l|l|l|}
\hline
Parameter & $R^2$ & RMSE & PCC & $R_{max}$ & KL-score ($\times 10^{-3}$) & JS-score ($\times 10^{-3}$) \\
\hline
$N=0$ & -0.337 & 54.948 & 0.007 & 200  & 169.917 & 41.288 \\
$N=1$ & 0.476  & 34.394 & 0.734 & 100  & 60.467  & 15.405 \\
$N=2$ & 0.863  & 17.620 & 0.933 & 50   & 21.176  & 5.210 \\
$N=3$ & 0.963  & 9.177  & 0.982 & 25   & 5.978   & 1.477 \\
$N=4$ & 0.990  & 4.720  & 0.995 & 12.5 & 1.638   & 0.408 \\
$N=5$ & 0.998  & 2.366  & 0.999 & 6.25 & 0.412   & 0.103 \\
\hline
\end{tabular}
\end{table}

In summary, in this paper, we mainly explored using these metrics to conduct some research and analysis on data flow within the system. Based on the similarities and differences shown by these metrics in the experimental results, we believe that in the system's internal evaluation, PCC and $R_{max}$ can be adopted because they provide a valuable reference for us to analyze whether the predicted positions and the ground-truth positions satisfy linearity and robustness as $N$ increases in Fig. \ref{fig: scatter}. Additionally, PCC also serves as a backup for $R^2$'s function. 
For external evaluation of system output, the combination of $R^2$ and RMSE allows us to focus more on improving model performance, as evaluated in the \cite{feng2018neural,feng2020weakly,feng2021vif,Glaser2020machine}.

\subsection{The qualitative metrics of the system in mathematics and statistics}
\label{xy-positions}
\subsubsection{x-position}

\begin{figure}   
    \centering
    \includegraphics[width=\textwidth]{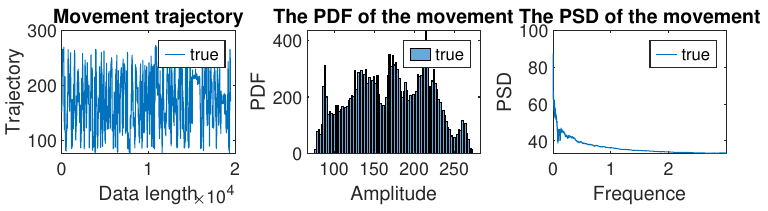}   
    \caption{From left to right, the actual waveform, the histogram of probability density function (PDF), and power spectral density (PSD) of the ground-truth moving positions are shown in the $x$-position. Among them, its PDF does not have significant distribution characteristics, such as non-Gaussian distribution, while its PSD decreases gradually and tends to a constant.
    }
    \label{fig: PDF-movement-x}
\end{figure}

In this section, we evaluate decoding predictions using probability density function (PDF) and power spectral density (PSD) in the context of internal correction within the system, observing what happens in neural data processing from the perspective of data distribution. 
Among them, PDF is adopted to evaluate whether the density distribution of the predicted positions matches the ground-truth positions distribution, as well as the noise distribution between them. PSD is adopted to evaluate the energy distribution of the predicted and ground-truth positions, and noise to observe changes in the system during data processing. 
In addition, in depicting the data distribution, we mainly focus on the quantitative indicators of KL-divergence and JS-divergence as referenced in Figs. \ref{fig: N-value-x} and \ref{fig: N-value-y}. 
Firstly, Fig. \ref{fig: PDF-movement-x} gives a visualization of the ground-truth trajectory in the $x$-position and its probability density distribution and power spectral density. It can be observed from the PDF and PSD that the PDF of the moving positions in the $x$-position does not have a distinct fixed distribution, while the PSD of the moving positions gradually decreases and stabilizes at a constant.

\begin{figure}    
    \centering
    \subfigure[The evaluation of the prediction in $x$-position (N=0,1,2)]{
        \includegraphics[width=0.47\textwidth]{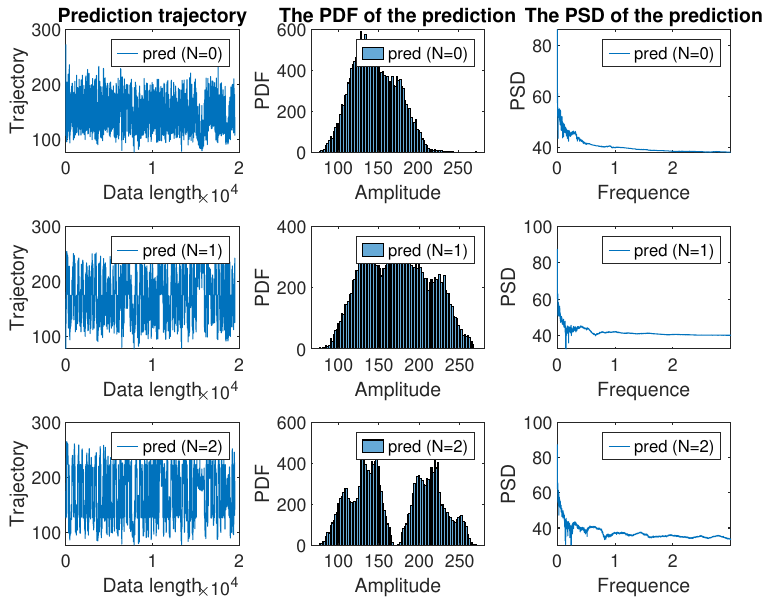}
    }
    \subfigure[The evaluation of the prediction in $x$-position (N=3,4,5)]{
        \includegraphics[width=0.47\textwidth]{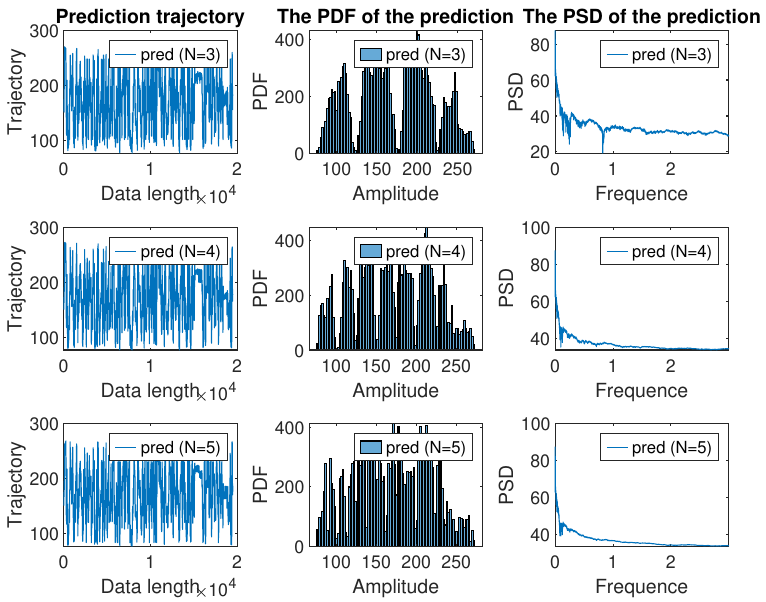}
    }
    \caption{(a) and (b) illustrate the waveform, PDF and PSD of the predicted positions in the $x$-positions from $N=0$ to $N=2$ and from $N=3$ to $N=5$, respectively.
    }
    \label{fig: PDF-prediction-x}
\end{figure}

\begin{figure}    
    \centering
    \subfigure[The evaluation of the noise in $x$-position (N=0,1,2)]{
        \includegraphics[width=0.47\textwidth]{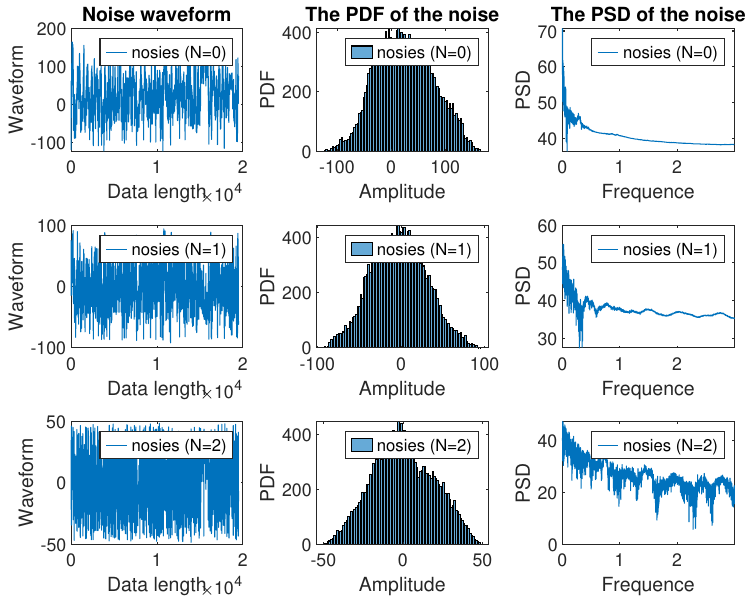}
    }
    \subfigure[The evaluation of the noise in $x$-position (N=3,4,5)]{
        \includegraphics[width=0.47\textwidth]{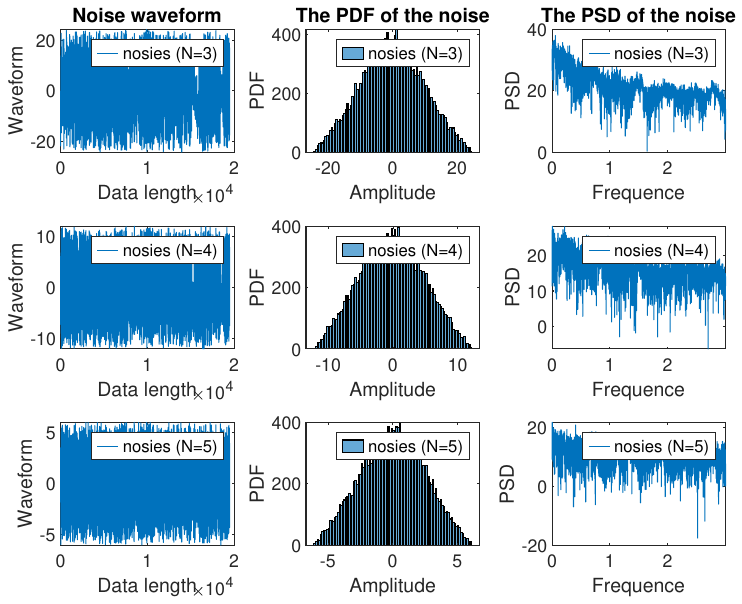}
    }
    \caption{(a) and (b) illustrate the waveform, PDF and PSD of the prediction error (i.e., noise) in the $x$-positions from $N=0$ to $N=2$ and from $N=3$ to $N=5$, respectively.
    }
    \label{fig: PDF-noise-x}
\end{figure}

Fig. \ref{fig: PDF-prediction-x} shows the PDF and PSD of the unsupervised predicted positions ($N=0$) and the corrected predicted positions ($N>0$) in $x$-position. When $N=0$, it can be observed that its PDF is close to a Gaussian distribution. However, after one correction ($N=1$), the width range of its distribution increases on the amplitude axis and the height range decreases from about 600 to about 400, but still maintains a Gaussian-like distribution. 
Then, when $N=2$, a significant change occurs in the PDF of the predicted positions, with two Gaussian-like distributions emerging. Similarly, it is observed that when $N=3$, four Gaussian-like distributions emerge; when $N=4$, eight Gaussian-like distributions emerge; when $N=5$, sixteen Gaussian-like distributions emerge, with $2^N$. 
At this point, when $N=5$, the PDF of the predicted positions is very similar to that of the ground-truth positions shown in Fig. \ref{fig: PDF-movement-x}. 
We believe that this phenomenon implies that as the $N$-value increases, the PDF of predicted positions approaches that of the ground-truth positions in the form of multiple Gaussian-like distributions. 
It is extremely important that as shown in Fig. \ref{fig: PDF-prediction-x}, the data distribution for the predicted and ground-truth positions gradually converges as the $N$-value increases, consistent with the evaluation of quantitative indicators (i.e., KL-divergence and JS-divergence values tend towards 0) in Table \ref{fig: N-value-x}. 
Finally, in PSD of the predicted positions, as $N$-value increases, it remains basically stable at about 40 and does not show significant changes.

Fig. \ref{fig: PDF-noise-x} demonstrates the PDF and PSD of the prediction error (i.e., noise) between the predicted positions and the ground-truth positions in the $x$-position. From the figure, it can be seen that as $N$-value increases, regardless of how the predicted position distribution changes in Fig. \ref{fig: PDF-prediction-x}, the prediction error always satisfies a Gaussian distribution of noise. This is a very obvious characteristic in Fig. \ref{fig: PDF-noise-x}. 
In addition, from the noise waveform, as $N$-value increases, it always floats around the zero-value, i.e., its mean is zero. On the other hand, the average value of its PDF can also be observed to be approximately zero-value.
However, unlike the PSD of the predicted and ground-truth positions, as $N$-value increases, the PSD of noise will also basically stabilize at a constant value, but there will be more spikes. At present, we do not know the specific role of these spikes and how they are generated, but we believe that they are likely to be a side effect generated to support prediction errors or noise that meet a Gaussian distribution.

\subsubsection{y-position}

\begin{figure}   
    \centering
    \includegraphics[width=\textwidth]{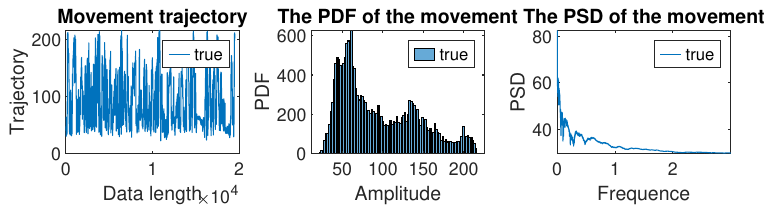}   
    \caption{From left to right, the actual waveform, the histogram of probability density function (PDF), and power spectral density (PSD) of the ground-truth moving positions are shown in the $y$-position. 
    }
    \label{fig: PDF-movement-y}
\end{figure}

In addition to testing on the $x$-position, testing on the $y$-position is also given. Fig. \ref{fig: PDF-movement-y} gives a visualization of the ground-truth trajectory in $y$-position and its probability density distribution and power spectral density. It can be observed from the PDF and PSD that the PDF of the moving positions does not also have a distinct fixed distribution, while the PSD of the moving positions gradually decreases and stabilizes at a constant as that in the $x$-position. 

\begin{figure}    
    \centering
    \subfigure[The evaluation of the prediction in $y$-position (N=0,1,2)]{
        \includegraphics[width=0.47\textwidth]{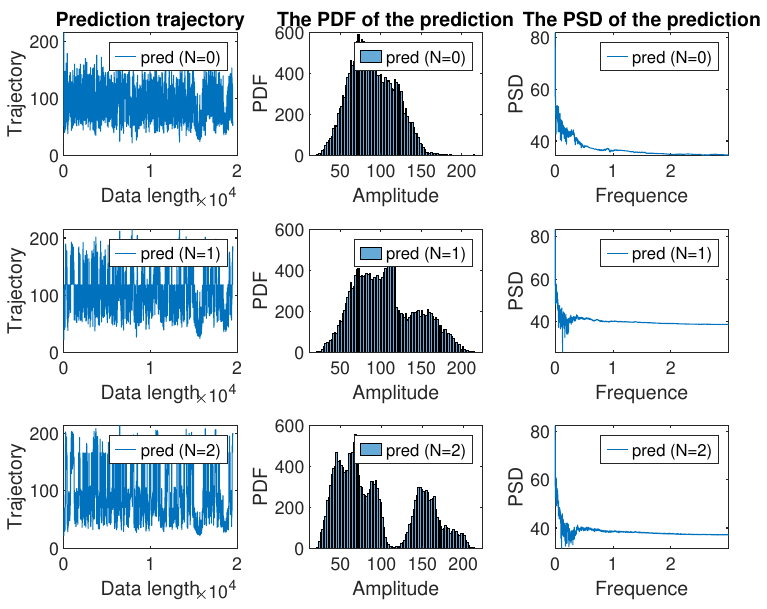}
    }
    \subfigure[The evaluation of the prediction in $y$-position (N=3,4,5)]{
        \includegraphics[width=0.47\textwidth]{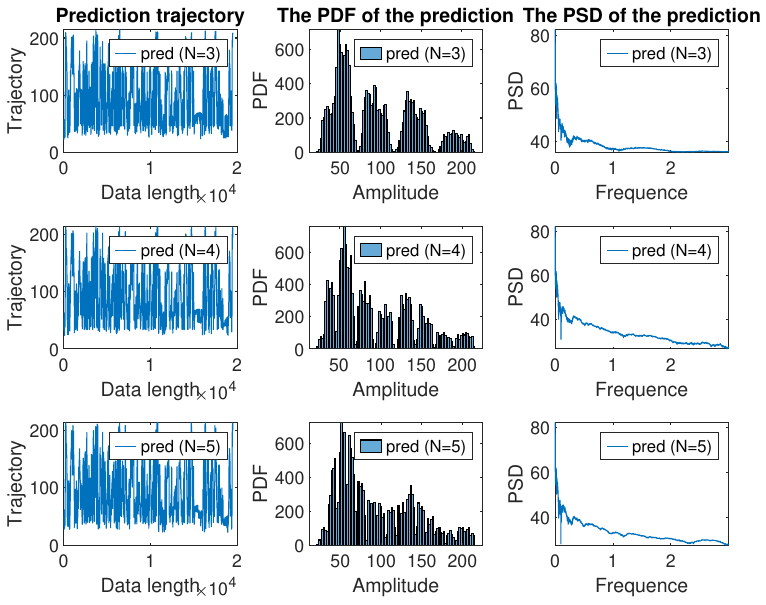}
    }
    \caption{(a) and (b) illustrate the waveform, PDF and PSD of the predicted positions in the $y$-positions from $N=0$ to $N=2$ and from $N=3$ to $N=5$, respectively.
    }
    \label{fig: PDF-prediction-y}
\end{figure}

\begin{figure}    
    \centering
    \subfigure[The evaluation of the noise in $y$-position (N=0,1,2)]{
        \includegraphics[width=0.47\textwidth]{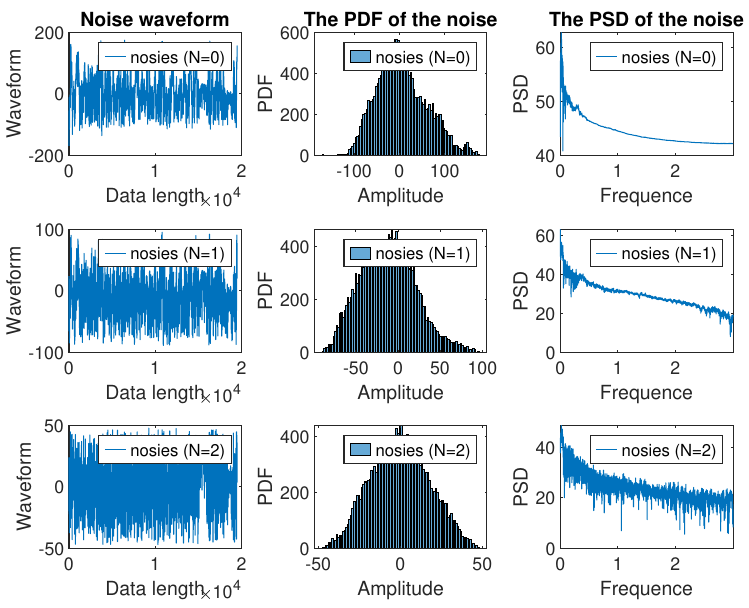}
    }
    \subfigure[The evaluation of the noise in $y$-position (N=3,4,5)]{
        \includegraphics[width=0.47\textwidth]{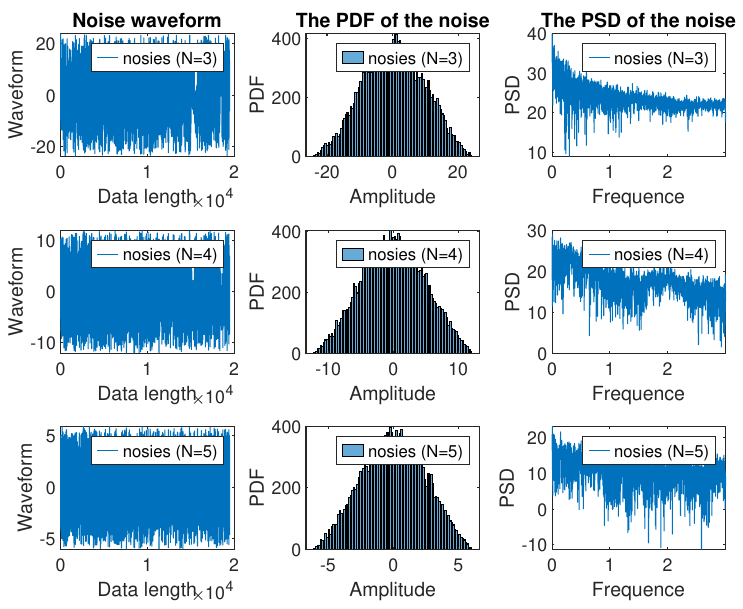}
    }
    \caption{(a) and (b) illustrate the waveform, PDF and PSD of the prediction error (i.e., noise) in the $y$-positions from $N=0$ to $N=2$ and from $N=3$ to $N=5$, respectively.
    }
    \label{fig: PDF-noise-y}
\end{figure}

Further, Fig. \ref{fig: PDF-prediction-y} shows the PDF and PSD of the unsupervised predicted positions ($N=0$) and the corrected predicted positions ($N>0$) in the $y$-position. As shown in the figure, as the $N$-value increases, the PDF of predicted positions also exhibits the characteristic of being composed of multiple Gaussian-like distributions, with several Gaussian-like distributions of $2^N$. 
One particularly significant feature is that while the PDF of the unsupervised prediction positions is Gaussian distribution when $N=0$, the PDF of the predicted positions is substantially consistent with that of the ground-truth positions in Fig. \ref{fig: PDF-movement-y} when $N=5$. Then, the PSD of the predicted positions gradually decreases and tends towards a constant. 
Finally, Fig. \ref{fig: PDF-noise-y} demonstrates the PDF and PSD of the prediction error (i.e., noise) between the predicted positions and the ground-truth positions in the $y$-position. Here, as can be seen from the graph, the PDF of the prediction error still satisfies a Gaussian distribution as the $N$-value increases, similar to the noise distribution in Fig. \ref{fig: PDF-noise-x}. 
Of course, just as the noise's PDF on the $x$-position, the noise's PDF on the $y$-position also exhibits more spikes as the $N$-value increases.

In summary, it deserves to be emphasized that in reference \cite{feng2020weakly}, weakly-supervised methods mainly adopted unsupervised EM and unsupervised KF as the basis for testing, and the decoding effect of EM was superior to that of the KF algorithm.
In this paper, we demonstrated the decoding prediction and data processing within the system using unsupervised EM (Un-EM) in reference \cite{feng2021vif}. Nevertheless, after previous preliminary testing, the decoding prediction and data processing of unsupervised KF \cite{feng2020weakly} still possess all the characteristics analyzed above for unsupervised EM within the system.
We believe that unsupervised methods should have mined inherent patterns possessed by this brain neural data (such as the brain area for decoding movement positions), namely symmetry and Gaussian distribution. Finally, regarding the consistency of data distribution, we think that KL-divergence and JS-divergence can be utilized to assess the distribution of data (i.e., between the predicted trajectory and the ground-truth trajectory) within the system.

\section{An algorithm board derived from the discovered symmetry}
\label{algorithm board}

\begin{figure}   
    \centering
    \includegraphics[width=\textwidth]{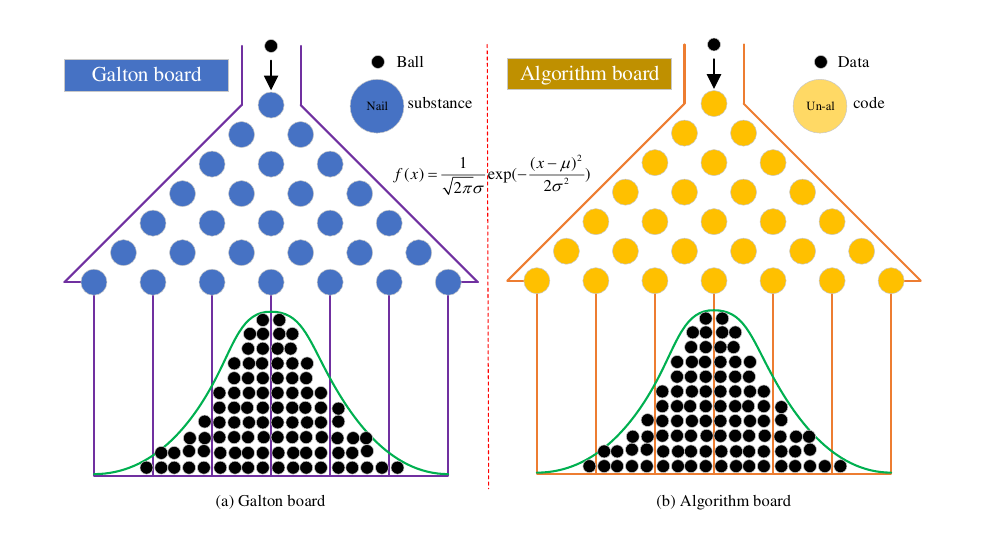}   
    \caption{A Galton board and an algorithm board.
    }
    \label{fig: Grid-SD2E_board}
\end{figure}

As observed from the aforementioned experiments and analysis and Fig. \ref{fig: correction}, the processing steps between the unsupervised method and the Galton board can be analogized as shown in Fig. \ref{fig: Grid-SD2E_board}.
Fig. \ref{fig: Grid-SD2E_board} (a) illustrates how a large number of balls traverse the Galton board in a binomial distribution, resulting in a normal distribution that can be proven using the central limit theorem.
Fig. \ref{fig: Grid-SD2E_board} (b) displays how a large amount of data undergoes unsupervised decoding, leading to a similar binomial distribution and, thus, a decoded position that can form a normal distribution. We believe that there is a strong similarity between them. If this is the case, when processing neural data, the brain should process in a manner similar to the processing pattern of the Galton board.
If this observation proves true, it will inspire us to develop new algorithms that are more similar to the brain's cognitive function system, that is, neural encoding and neural decoding.

\section{Conclusion}
\label{conclusion}

In this paper, we explore the information contained in the discovered symmetry from various indicators. As a conclusion, the overall perspective is as follows:
(1) The probability density function (PDF) of the prediction location of the unsupervised method follows a Gaussian distribution. Moreover, noise, defined as the difference between the ground-truth location and the predicted location, has a PDF that always follows a Gaussian distribution.
(2) As the $N$-value increases, the PDF of the prediction location gradually approaches the PDF of the ground-truth location  (such as, non-Gaussian distribution, etc.) using multiple Gaussian distributions.
(3) As the $N$-value increases, whether it is the ground-truth location, predicted location, or the noise between them, their power spectral density (PSD) will gradually tend towards a constant; however, the PSD of noise will have more spikes.
Lastly, an algorithm board similar to the Galton board was constructed to serve as the mathematical foundation of the discovered symmetry.

%
%
%
\bibliographystyle{splncs04}
%

\end{document}